\documentclass[runningheads]{llncs}
\usepackage{makecell}
\usepackage{multirow}
\usepackage{adjustbox}
\usepackage{subcaption}
\usepackage{amsmath}
\usepackage{amssymb}
\usepackage{marvosym} 
\usepackage[T1]{fontenc}
\usepackage{graphicx,verbatim}
\begin{document}
\title{LGE-Guided Cross-Modality Contrastive Learning for Gadolinium-Free Cardiomyopathy Screening in Cine CMR}
%

\author{Siqing Yuan\inst{1, 2} \and Yulin Wang\inst{1} \and Zirui Cao\inst{2} \and Yueyan Wang\inst{2} \and Zehao Weng\inst{1, 2} \and \\Hui Wang\inst{4} \and Lei Xu\inst{4} \and Zixian Chen\inst{5} \and Lei Chen\inst{2}\textsuperscript{(\Letter)} \and Zhong Xue\inst{2} \and \\
Dinggang Shen\inst{1, 2, 3}\textsuperscript{(\Letter)}}  
\institute{  
  School of Biomedical Engineering and State Key Laboratory of Advanced Medical Materials and Devices, ShanghaiTech University, Shanghai 201210, China \\
  \email{dinggang.shen@gmail.com}  
  \and  
  Department of Research and Development, United Imaging Intelligence, Shanghai 200230, China \\
  \email{lei.chen@uii-ai.com}  
  \and  
  Shanghai Clinical Research and Trial Center, Shanghai 201210, China
  \and  
  Department of Radiology, Beijing Anzhen Hospital, Capital Medical University, Beijing 100029, China
  \and  
  Department of Radiology, The First Hospital of Lanzhou University, Gansu 730000, China
}

\maketitle              
\begin{abstract}

Cardiomyopathy, a principal contributor to heart failure and sudden cardiac mortality, demands precise early screening. Cardiac Magnetic Resonance (CMR), recognized as the diagnostic ``gold standard'' through multiparametric protocols, holds the potential to serve as an accurate screening tool. However, its reliance on gadolinium contrast and labor-intensive interpretation hinders population-scale deployment. We propose \textbf{CC-CMR}, a \textbf{C}ontrastive Learning and \textbf{C}ross-Modal alignment framework for gadolinium-free cardiomyopathy screening using cine \textbf{CMR} sequences. By aligning the latent spaces of cine CMR and Late Gadolinium Enhancement (LGE) sequences, our model encodes fibrosis-specific pathology into cine CMR embeddings. A Feature Interaction Module concurrently optimizes diagnostic precision and cross-modal feature congruence, augmented by an uncertainty-guided adaptive training mechanism that dynamically calibrates task-specific objectives to ensure model generalizability. Evaluated on multi-center data from 231 subjects, CC-CMR achieves accuracy of 0.943 (95\% CI: 0.886–0.986), outperforming state-of-the-art cine-CMR-only models by 4.3\% while eliminating gadolinium dependency, demonstrating its clinical viability for wide range of populations and healthcare environments.


\end{abstract}

\section{Introduction}
Cardiomyopathy, characterized by structural and functional myocardial abnormalities, remains a principal etiology of heart failure and arrhythmias~\cite{2023ESC}. This disorder affects individuals across all age strata, with some cases exhibiting genetic inheritance (e.g., hypertrophic cardiomyopathy). Early screening for cardiomyopathy improves prognosis and identifies genetic risks in families, enabling timely intervention and personalized treatment.  

Cardiac Magnetic Resonance (CMR) imaging provides comprehensive myocardial functional and tissue characterization through multiparametric protocols, making it a diagnostic gold standard for cardiomyopathy evaluation~\cite{2023ESC}. Routine protocols include short-axis (SAX) and 4-chamber (4CH) cine CMR sequences for quantitative cardiac motion analysis, as well as SAX Late Gadolinium-Enhanced (LGE) sequences for identifying myocardial fibrosis and scar tissue (Fig. 1). Despite the diagnostic superiority of CMR, its clinical implementation is limited by (1) time-consuming cross-sequence analysis with observer variability and (2) restricted gadolinium (Gd) use in patients with renal impairment or pregnancy. These barriers hinder its scalability for population-level screening, highlighting the necessity of developing an automated, non-contrast CMR-based cardiomyopathy screening model. 

\begin{figure}[htbp]  
\centering 
\includegraphics[width=0.5\textwidth]{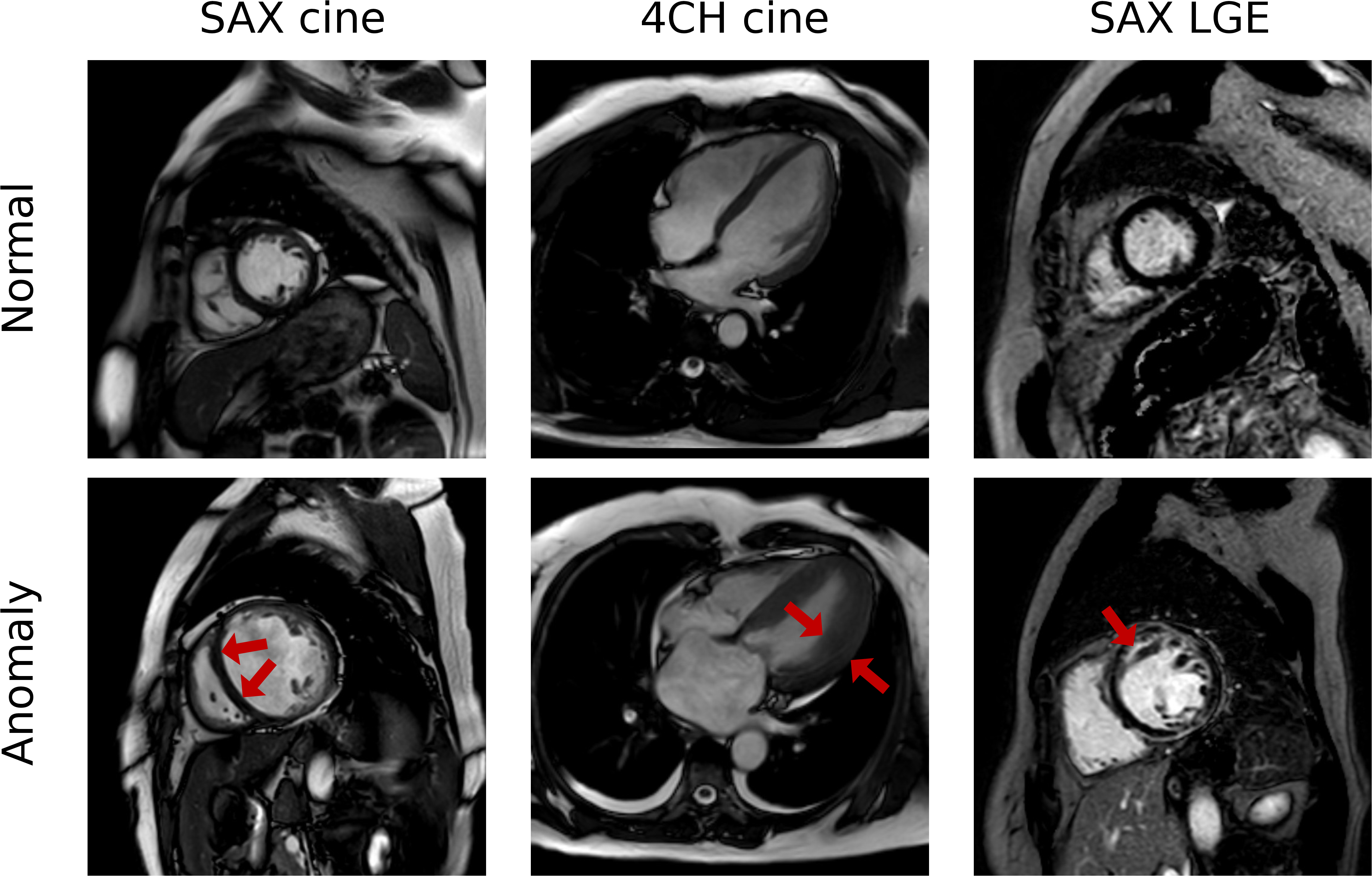}
\caption{Typical CMR findings in cardiomyopathy patients across different sequences, such as abnormal ventricular dilation (left), abnormal ventricular hypertrophy (middle), and abnormal hyperintense regions in myocardial tissue (right).} \label{fig1}
\end{figure}  









\textbf{Multimodal Learning} can integrate heterogeneous data to improve task performance, which matches the characteristics of multiple imaging sequences in CMR. Wang et al.~\cite{NatureMedicine} proposed a dual-view cine fusion framework that achieved state-of-the-art (SOTA) results in Cardiovascular Disease (CVD) screening. However, it excludes the available LGE data which are routinely acquired in current clinical protocols. Given the unique advantage of LGE sequences in identifying myocardial fibrosis and scar, excluding them from analysis undermines the capability of CMR in cardiomyopathy detection. Zhu et al.~\cite{ECG} improved CVD detection performance by integrating multimodal features at multiple scales. However, this strict multimodal integration approach fails to accommodate missing input modalities during inference, limiting its applicability when excluding LGE sequences in the inference phase.




\textbf{Contrastive Learning} provides a promising solution by aligning cross-modal representations in shared embedding spaces, enabling robust knowledge transfer across modalities. The seminal Contrastive Language-Image Pretraining (CLIP)~\cite{CLIP} pioneered the application paradigm of this approach, and recent adaptations in medical imaging show its significant promise. For example, Yu et al.~\cite{CoreCLIP} developed a core-periphery feature alignment approach between medical images and clinical text for zero-shot diagnostic transfer, and Sun et al.~\cite{NoduleCLIP} achieved SOTA pulmonary nodule classification results by aligning image features and class features through their Nodule-CLIP framework. These advancements demonstrate that cross-modal alignment can effectively preserve diagnostic semantics through modality-invariant feature learning, thereby reducing reliance on specific imaging modalities while maintaining accuracy.




Building upon these insights, we propose \textbf{CC-CMR}, a \textbf{C}ontrastive learning-based framework that incorporates \textbf{C}ross-modal alignment into multimodal \textbf{CMR} representation models. Our architecture implicitly encodes pathological information specific to LGE sequences into the representation of cine CMR sequences, leveraging complementary information to improve screening (a binary classification task to determine whether a patient has cardiomyopathy) performance using only non-contrast sequences during inference. The core contributions of this study include:

\begin{enumerate}
  \item Provide a cardiomyopathy screening model that integrates pathological knowledge, achieving superior diagnostic performance with non-contrast cine CMR sequences, making it practical for primary care and health check-up centers.
  \item Propose a feature interaction module that simultaneously handles classification and feature alignment, introducing a novel extensible paradigm for multi-sequence analysis.
  \item Introduce an uncertainty-aware mechanism that balances various objectives in the feature interaction module, enhancing the generalizability of the model.
\end{enumerate}

Our method achieves higher classification accuracy (94.3\% vs. 90.0\%) and F1-score (0.946 vs. 0.907) compared to the SOTA model~\cite{NatureMedicine}, demonstrating its potential for advanced cardiomyopathy screening.

\section{Method}
\subsection{Model Architecture}

\begin{figure}  
\includegraphics[width=\textwidth]{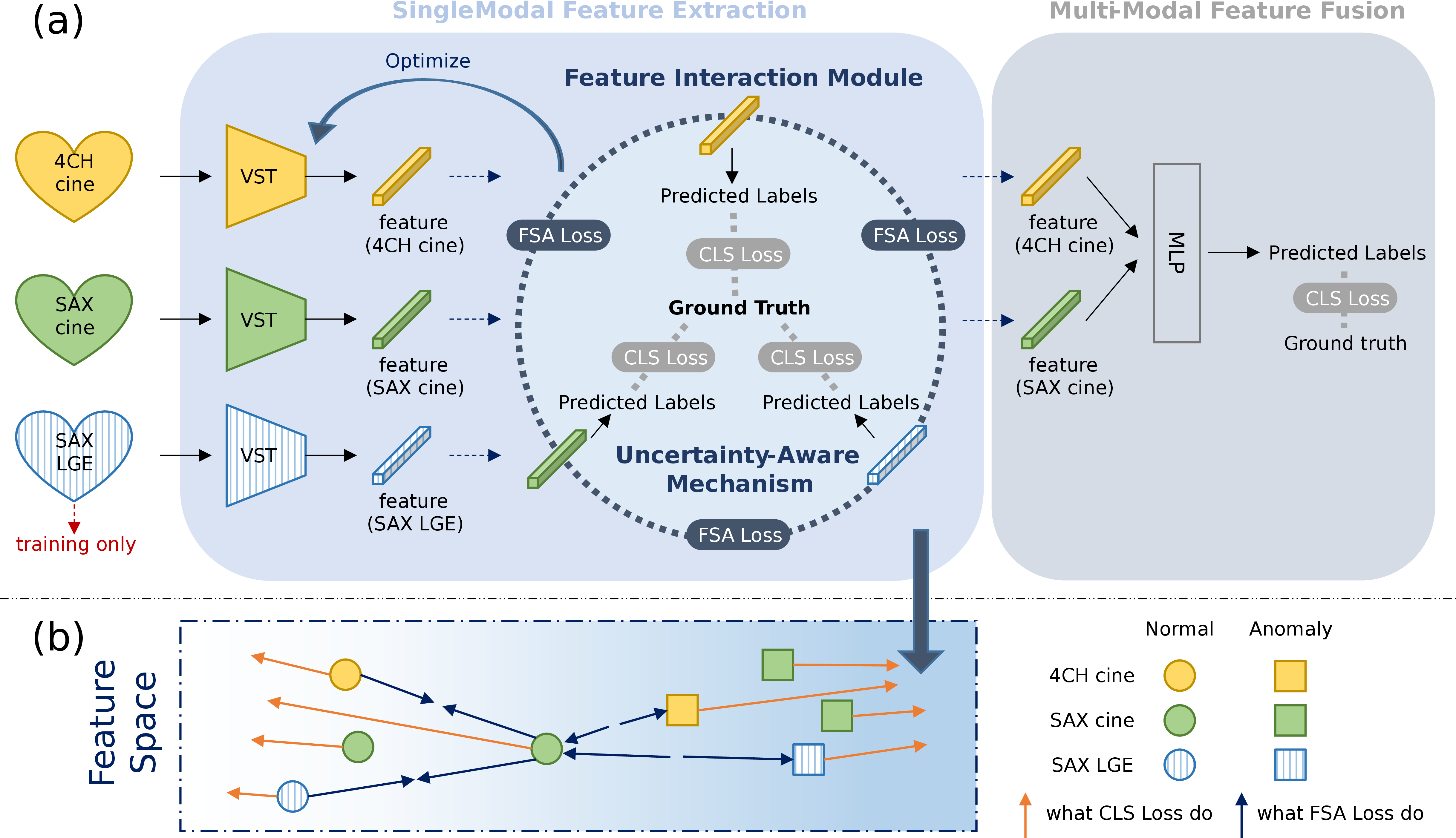}  
\caption{An overview of proposed CC-CMR, including (a) model structure and (b) schematic representation for the Feature Interaction Module.} \label{fig1}  
\end{figure}

Our framework aims to train a classification network that enables cine CMR sequences to learn pathological features visualized on LGE sequences. It comprises two key stages (as shown in Fig. 2(a)): 1) Single-Sequence Feature Extraction and 2) Multi-Sequence Feature Fusion, which are detailed below.

\textbf{Stage 1: Single-Sequence Feature Extraction.} Here, an innovative Feature Interaction Module is embedded to balance sequence-specific preservation and cross-sequence knowledge sharing. In particular, each sequence (4CH cine, SAX cine, SAX LGE) is encoded by a Video Swin Transformer (VST~\cite{VST}) pretrained on the Kinetics600 dataset, which has been proved efficient in CMR images feature extraction~\cite{NatureMedicine}. During training, the encoded features are processed by the \textbf{Feature Interaction Module} (detailed in Section 2.2). This module facilitates cross-sequence feature alignment and classification, enabling the cine CMR representations to implicitly incorporate LGE-related pathological patterns. Specifically, this module introduces two complementary loss functions: a classification loss (\textbf{CLS loss}) to ensure classification accuracy and a feature space alignment loss (\textbf{FSA loss}) to align features across sequences. An \textbf{Uncertainty-Aware Mechanism} (detailed in Section 2.3), embedded in the Feature Interaction Module, automatically balances the contributions of these two losses.

\textbf{Stage 2: Multi-Sequences Feature Fusion.} Aligned features from the two cine CMR sequences (4CH cine and SAX cine) are fed into a lightweight two-layer perceptron (MLP). This module integrates complementary information from multiple cine CMR views, generating the final classification output with optimized accuracy through CLS loss minimization.

\textbf{Training and Inference.} During training, all three sequences (4CH cine, SAX cine, and SAX LGE) are utilized to jointly refine the network, where LGE provides additional supervision to help the cine CMR extractors to capture fine-grained pathological patterns crucial for cardiac diagnosis. During inference, only cine CMR sequences (4CH cine and SAX cine) are required for prediction, ensuring clinical feasibility by avoiding reliance on gadolinium-based inputs.

\subsection{Feature Interaction Module}

This module is a key component of our framework, used to enable the entire model to function effectively. To facilitate effective cross-sequence learning while preserving discriminative power, we propose a dual-loss optimization strategy consisting of CLS loss and FSA loss. By synergistically guiding feature distributions across sequences, these two losses collectively drive the embedding features toward an optimal shared space that integrates discriminative characteristics from all sequences (Fig. 2(b)).


\subsubsection{Classification Loss} guide the feature vectors of each sequence toward a more distinguishable direction in feature space (Fig. 2(b)), thereby preserving the specific classification information. We employ the focal loss proposed by He et al.~\cite{Focalloss}. For a batch with $N$ samples in sequence $I$, the CLS loss is calculated as the average focal loss across these samples, as given by the following equation:
\begin{equation}  
\mathcal{L}_{\mathrm{CLS}}\left(I\right) = -\frac{1}{N} \sum_{i=1}^{N}\left(1-p_{i}\right)^{\gamma} \log \left(p_{i}\right),
\end{equation} 
where $p_{i}$ is specifically defined as the predicted probability assigned to the true label of the sample. As $p_{i} \to 1$, the modulating factor $\left(1-p_{i}\right)^{\gamma} \to 0$, reducing the contribution of such well-predicted samples to the loss, thus emphasizing the training on hard-to-distinguish cases. The focusing parameter $\gamma$ is set to $2$ by following the optimal setting from~\cite{Focalloss}, which controls the degree of loss attenuation.

\subsubsection{Feature Space Alignment Loss} encourages embeddings of the same class from different sequences to converge in the feature space while driving those belonging to different classes further apart. (Fig. 2(b)). We define FSA loss through an adapted implementation of InfoNCE loss in work~\cite{CLIP}, which enforces semantic consistency by maximizing cosine similarity between positive pairs (same instances across sequences) and repelling negative pairs (different instances across sequences)~\cite{preCLIP}.

For a batch containing $N$ instances, each instance should correspond to samples from three sequences. For any two sequences $I$ and $J$, let $\vec{v}_{i}$ and $\vec{u}_{j}$ represent the feature vectors of sample $i$ from sequence $I$ and sample $j$ from sequence $J$, respectively. The FSA loss between them is defined as follows:
\begin{equation}
\mathcal{L}_{\mathrm{FSA}}\left(I,J\right) = 
-\frac{1}{2 N} \sum_{i=1}^{N} \sum_{j=1}^{N}
y_{i j} \log \frac{e^{\text{sim}(\vec{v}_{i},\vec{u}_{j})}}{\sum_{k=1}^{N} e^{\text{sim}(\vec{v}_{i},\vec{u}_{k})}}
+y_{j i} \log \frac{e^{\text{sim}(\vec{v}_{i},\vec{u}_{j})}}{\sum_{k=1}^{N} e^{\text{sim}(\vec{v}_{k},\vec{u}_{j})}},
\end{equation}
where the indicator $y_{i j}=1$ if $i$ and $j$ are from the same patient, otherwise $y_{i j}=0$. $\text{sim}(\vec{u}_{i},\vec{v}_{j})$ calculates the cosine similarity between $\vec{u}_{i}$ and $\vec{v}_{j}$:
\begin{equation}
\text{sim}\left(\vec{u}_{i}, \vec{v}_{j}\right)=\frac{\vec{u}_{i} \cdot \vec{v}_{j}}{\left\|\vec{u}_{i}\right\| \left\|\vec{v}_{j}\right\|}.
\end{equation}

While sample-level alignment enforces strict pairwise feature matching, it may suppress sequence-specific discriminative features. Therefore, in practical implementation, we first merge and average the cosine similarity matrices of the samples based on their labels, resulting in a label-based cosine similarity matrix for subsequent calculation. In this context, $i$ and $j$ in the formula actually represent the label indices.

\subsection{Uncertainty-Aware Mechanism}

The conflicting objectives of CLS loss (preserving intra-sequence discriminability) and FSA loss (enforcing inter-sequence consistency) induce gradient conflicts during joint optimization. Manual weight tuning requires exhaustive trial-and-error iterations and fails to accommodate phase-dependent dynamics, often leading to suboptimal trade-offs: (1) insufficient alignment, leading to sequence-specific feature drift, or (2) excessive alignment, compromising intra-sequence classification separability.

To address this, we introduce an Uncertainty-Aware Mechanism into the Feature Interaction Module by incorporating uncertainty parameters $\sigma$. These parameters are modeled as \textbf{learnable variables} reflecting the Homoscedastic uncertainty of each task~\cite{Uncertainty}~\cite{Uncertainty2}:
\begin{equation}
\mathcal{L} = 
\sum \frac{1}{\sigma_{I J}^{2}} \mathcal{L}_{\mathrm{FSA}}\left(I,J\right) 
+ \sum \frac{1}{\sigma_{I}^{2}} \mathcal{L}_{\mathrm{CLS}}\left(I\right)
+ \sum \log \left( \sigma_{I J} + 1 \right) 
+ \sum \log \left( \sigma_{I} + 1\right),
\end{equation}
where $\sigma_{I}$ captures classification uncertainty in sequence $I$, while $\sigma_{I J}$ measures inter-sequence alignment uncertainty between sequences $I$ and $J$. During optimization, the weights inversely scale with task uncertainty $\sigma$, with higher uncertainty suppressing the loss influence to resolve gradient conflicts. The logarithmic regularization term $\log \left( \sigma + 1\right)$ imposes a penalty on large uncertainty parameters ($\sigma \to \infty$), which could lead to a corresponding weight of zero. This mechanism enables stable co-optimization, balancing intra-sequence discriminative power with cross-sequence representation consistency.


\section{Experiments}
\subsection{Implementation Details}
The study cohort comprised 231 subjects (109 healthy controls and 122 patients) from Beijing Anzhen Hospital and The First Hospital of Lanzhou University: the testing subset consisting of 33 healthy controls and 37 patients, while the training subset including 76 healthy controls and 85 patients. The training set was augmented with four-fold resampling. This cohort covers all cardiomyopathy subtypes defined by the ESC 2023 guidelines~\cite{2023ESC}. Multi-sequence CMR acquisitions were performed using Philips, GE, and Siemens scanners. We typically select: one mid-ventricular slice from the 4CH cine, mid-3 slices from the SAX cine, and mid-6 slices from the SAX LGE series.

The 3D model, tiny VST, serves as the feature extractor. Input dimensions for cine images are (batch size, number of slices, number of frames, 224, 224), while for LGE images, they are (batch size, RGB channels duplicated from intensity, number of slices, 224, 224). In-plane images are resized to 224 × 224 pixels, with a pixel size of 0.994 mm × 0.994 mm. Pixel values were truncated at the 0.1th and 99.9th percentiles, followed by Z-score intensity normalization.

All experiments were conducted on NVIDIA L40 GPUs through \textbf{three training phases}: (i) Single-sequence feature extractor pretraining (300 epochs per sequence with classification loss, initialized from Kinetics600-pretrained VST); (ii) Single-sequence feature extractor finetuning (600 epochs with Feature Interaction Module); (iii) Feature fusion network training (10 epochs with frozen VST). The AdamW optimizer, with an initial learning rate of 0.001 and cosine annealing for learning rate scheduling, was employed across all phases.

\subsection{Experiments Results}
\subsubsection{Performance Comparison.}
Table 1 compares our proposed CC-CMR model with the SOTA CVD screening model by Wang et al.~\cite{NatureMedicine} in distinguishing healthy controls from cardiomyopathy patients. Efron's bootstrap method \cite{Boostrap} with 1,000 resamples was used to estimate confidence intervals. As shown, our model outperforms the SOTA method on all classification metrics, with accuracy increasing from 0.900 to 0.943 and a notable improvement in both sensitivity (+2.7\%) and specificity (+6.0\%), while maintaining a comparable AUC. These results demonstrate superior diagnostic accuracy under non-Gd conditions, and the narrower confidence intervals further indicate enhanced model stability.

\begin{table}
\caption{Performance comparison of CC-CMR vs. SOTA CVD screening model.}\label{tab1}
\centering
\adjustbox{width=\textwidth}{ 
\begin{tabular}{lccccc}
\hline
\multirow{2}{*}{Method} & \multicolumn{5}{c}{Metrics (CI: 95\%)} \\
\cline{2-6}
& Accuracy & F1-score & Sensitivity & Specificity & AUC \\
\hline

SOTA~\cite{NatureMedicine} & \makecell{0.900 \\ (0.829,0.957)} 
& \makecell{0.907 \\ (0.830,0.968)} 
& \makecell{0.919 \\ (0.881,0.971)} 
& \makecell{0.879 \\ (0.856,0.969)} 
& \makecell{0.955 \\ (0.906,0.992)} \\
CC-CMR & \makecell{\textbf{0.943} \\ (0.886,0.986)} 
& \makecell{\textbf{0.946} \\ (0.882,0.987)} 
& \makecell{\textbf{0.946} \\ (0.942,0.971)} 
& \makecell{\textbf{0.939} \\ (0.934,0.974)} 
& \makecell{\textbf{0.957} \\ (0.899,0.998)} \\
\hline
\end{tabular}
}
\end{table}


\subsubsection{Ablation Experiments.}

\begin{table}
\caption{Ablation results on input sequences and model components (FIM: Feature Interaction Module, U-A: Uncertainty-Aware Mechanism).}\label{tab1}
\centering
\adjustbox{width=\textwidth}{ 
\begin{tabular}{ccccccccc}
\hline
\multirow{2}{*}{\makecell{4CH \\ cine}}
& \multirow{2}{*}{\makecell{SAX \\ cine}}
& \multirow{2}{*}{\makecell{FIM \\ w/o U-A}}
& \multirow{2}{*}{\makecell{FIM \\ w U-A}}
& \multicolumn{5}{c}{Metrics (CI: 95\%)} \\
\cline{5-9}
& & & & Accuracy & F1-score & Sensitivity & Specificity & AUC \\
\hline

\checkmark & & & 
& \makecell{0.829 \\ (0.743,0.914)}
& \makecell{0.842 \\ (0.747,0.920)}
& \makecell{0.865 \\ (0.811,0.929)}
& \makecell{0.788 \\ (0.796,0.859)}
& \makecell{0.937 \\ (0.871,0.979)} \\
 & \checkmark & & 
& \makecell{0.800 \\ (0.700,0.886)}
& \makecell{0.806 \\ (0.690,0.893)}
& \makecell{0.784 \\ (0.813,0.841)}
& \makecell{0.818 \\ (0.786,0.876)}
& \makecell{0.909 \\ (0.823,0.972)} \\
\checkmark & \checkmark & & 
& \makecell{0.900 \\ (0.829,0.957)}
& \makecell{0.907 \\ (0.830,0.968)}
& \makecell{0.919 \\ (0.881,0.971)}
& \makecell{0.879 \\ (0.856,0.969)}
& \makecell{0.955 \\ (0.906,0.992)} \\
\checkmark & \checkmark & \checkmark & 
& \makecell{0.929 \\ (0.871,0.986)}
& \makecell{0.932 \\ (0.862,0.983)}
& \makecell{0.919 \\ (0.949,0.967)}
& \makecell{0.939 \\ (0.936,0.973)}
& \makecell{0.955 \\ (0.896,0.995)} \\
\checkmark & \checkmark & & \checkmark 
& \makecell{\textbf{0.943} \\ (0.886,0.986)}
& \makecell{\textbf{0.946} \\ (0.882,0.987)}
& \makecell{\textbf{0.946} \\ (0.942,0.971)}
& \makecell{\textbf{0.939} \\ (0.934,0.974)}
& \makecell{\textbf{0.957} \\ (0.899,0.998)} \\
\hline
\end{tabular}
}
\end{table}
Table 2 summarizes the classification performance under different input sequences and model components. Multi-sequence fusion (Row 3) outperforms the single-sequence inputs, with accuracy increases of 8.1\% and 10.0\% over Rows 1 and 2, respectively, demonstrating the substantial benefit of sequence integration. The Feature Interaction Module (Rows 4 and 5) further improves accuracy by 2.9\% and 4.3\% compared to fusion alone (Row 3), for Rows 4 and 5, respectively, illustrating the added value of feature space alignment. Additionally, the Uncertainty-Aware Mechanism raises accuracy by another 1.4 percentage points (Row 5 vs. Row 4), confirming the effectiveness of this strategy.


\subsubsection{Visualization of Feature Interaction Module Efficacy.}

\begin{figure}[htbp]  
    \centering  
    \begin{subfigure}{0.45\textwidth} 
        \centering  
        \includegraphics[width=\textwidth]{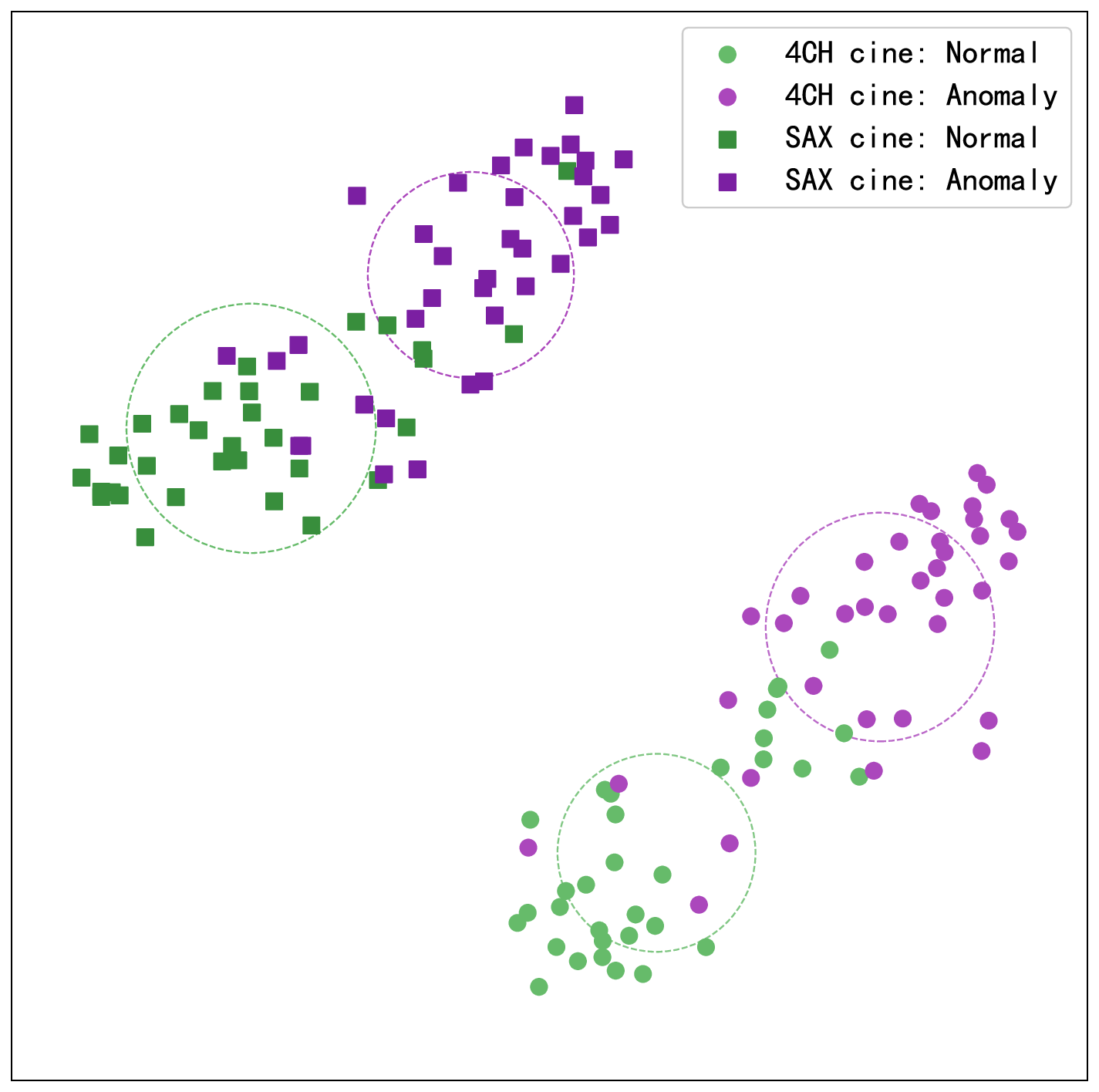}
        \caption{ w/o Feature Interaction Module}  
        \label{fig:subfig1}  
    \end{subfigure}   
    \begin{subfigure}{0.45\textwidth}  
        \centering  
        \includegraphics[width=\textwidth]{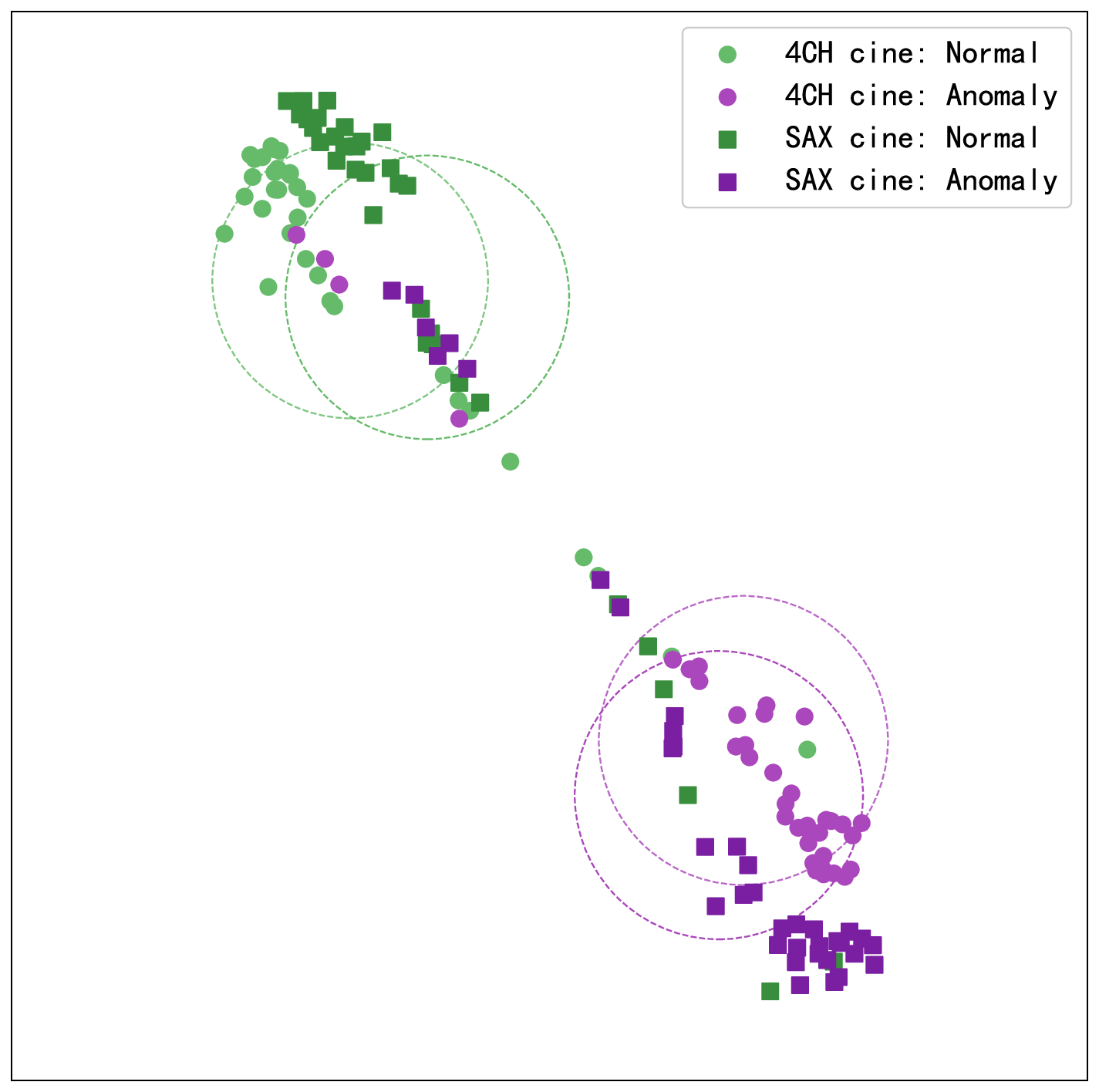}
        \caption{ w Feature Interaction Module}  
        \label{fig:subfig2}  
    \end{subfigure}  
    \caption{t-SNE visualization before and after using Feature Interaction Module.}
    \label{fig:overall}  
\end{figure}  

In Fig.~\ref{fig:overall}, t-SNE~\cite{tSNE} visualization illustrates the cross-sequence feature distributions of cine CMR sequences in the embedded space, with dashed contours indicating mean ± standard deviation ranges. After applying the Feature Interaction Module, the distributions between sequences become closer, while the relative distances between categories within each sequence increase. This suggests that the module can effectively align feature spaces while preserving vector separability.



To further test the performance of our framework on the subtyping of cardiomyopathy (Hypertrophic cardiomyopathy, Non-dilated left ventricular cardiomyopathy, Dilated cardiomyopathy, Restrictive cardiomyopathy and Arrhythmogenic right ventricular cardiomyopathy), we trained a five-group-classifier and compared it with the same SOTA model~\cite{NatureMedicine}. The results showed our model exhibits superior performance (Supplemental Material).

\section{Conclusion}


In this paper, we proposed CC-CMR, a contrastive learning-based framework for gadolinium-free cardiomyopathy screening. The advantage of our approach lies in the Feature Interaction Module with an Uncertainty-Aware Mechanism, which dynamically aligns LGE and cine CMR features during training and thereby enables the model to implicitly capture LGE-specific pathological patterns. Experimental results show that CC-CMR achieves higher accuracy for cardiomyopathy screening under non-contrast conditions. Moreover, the designed Feature Interaction Module is both flexible and extensible, enabling easy adaptation to additional imaging modalities. Extensive clinical validation will be conducted to further enhance the diagnostic capabilities and reliability of CC-CMR.

\begin{credits}
\subsubsection{\ackname}  This work was partially supported by the National Key Research and Development Program of China (2022YFE0209800).

\subsubsection{\discintname}
The authors have no competing interests to declare that are relevant to the content of this article.
\end{credits}


\bibliographystyle{splncs04}
\bibliography{reference}

\end{document}